\documentclass[letterpaper]{article} 
\usepackage{aaai25}  
\usepackage{times}  
\usepackage{helvet}  
\usepackage{courier}  
\usepackage[hyphens]{url}  
\usepackage{graphicx} 
\usepackage{natbib}  
\usepackage{caption} 
\usepackage{xcolor}
\usepackage{amssymb}
\usepackage{amsmath}
\usepackage{graphicx}
\usepackage{booktabs}
\usepackage{subcaption}
\usepackage{multirow}

\usepackage{multirow}
\usepackage{graphicx}
\usepackage{float}

\definecolor{White}{rgb}{1.,0.,1.}
\definecolor{first}{rgb}{.8,.0,.0}
\definecolor{second}{rgb}{.0,.6,.0}
\definecolor{third}{rgb}{.0,.0,.8}

\definecolor{ceiling}{RGB}{214,  38, 40}
\definecolor{floor}{RGB}{43, 160, 4}
\definecolor{wall}{RGB}{158, 216, 229}
\definecolor{window}{RGB}{114, 158, 206}
\definecolor{chair}{RGB}{204, 204, 91}
\definecolor{bed}{RGB}{255, 186, 119}
\definecolor{sofa}{RGB}{147, 102, 188}
\definecolor{table}{RGB}{30, 119, 181}
\definecolor{tvs}{RGB}{160, 188, 33}
\definecolor{furniture}{RGB}{255, 127, 12}
\definecolor{objects}{RGB}{196, 175, 214}

\definecolor{car}{rgb}{0.39215686, 0.58823529, 0.96078431}
\definecolor{bicycle}{rgb}{0.39215686, 0.90196078, 0.96078431}
\definecolor{motorcycle}{rgb}{0.11764706, 0.23529412, 0.58823529}
\definecolor{truck}{rgb}{0.31372549, 0.11764706, 0.70588235}
\definecolor{othervehicle}{rgb}{0.39215686, 0.31372549, 0.98039216}
\definecolor{person}{rgb}{1.        , 0.11764706, 0.11764706}
\definecolor{bicyclist}{rgb}{1.        , 0.15686275, 0.78431373}
\definecolor{motorcyclist}{rgb}{0.58823529, 0.11764706, 0.35294118}
\definecolor{road}{rgb}{1.        , 0.        , 1.        }
\definecolor{parking}{rgb}{1.        , 0.58823529, 1.        }
\definecolor{sidewalk}{rgb}{0.29411765, 0.        , 0.29411765}
\definecolor{otherground}{rgb}{0.68627451, 0.        , 0.29411765}
\definecolor{building}{rgb}{1.        , 0.78431373, 0.        }
\definecolor{fence}{rgb}{1.        , 0.47058824, 0.19607843}
\definecolor{vegetation}{rgb}{0.        , 0.68627451, 0.        }
\definecolor{trunk}{rgb}{0.52941176, 0.23529412, 0.        }
\definecolor{terrain}{rgb}{0.58823529, 0.94117647, 0.31372549}
\definecolor{pole}{rgb}{1.        , 0.94117647, 0.58823529}
\definecolor{trafficsign}{rgb}{1.        , 0.        , 0.        }
\definecolor{otherstructure}{rgb}{0.98039215, 0.58823529, 0.}
\definecolor{otherobject}{rgb}{0.19607843, 1.        , 1.        }

\makeatletter

\newcommand{\car@semkitfreq}{3.92}
\newcommand{\bicycle@semkitfreq}{0.03}
\newcommand{\motorcycle@semkitfreq}{0.03}
\newcommand{\truck@semkitfreq}{0.16}
\newcommand{\othervehicle@semkitfreq}{0.20}
\newcommand{\person@semkitfreq}{0.07}
\newcommand{\bicyclist@semkitfreq}{0.07}
\newcommand{\motorcyclist@semkitfreq}{0.05}
\newcommand{\road@semkitfreq}{15.30}
\newcommand{\parking@semkitfreq}{1.12}
\newcommand{\sidewalk@semkitfreq}{11.13}
\newcommand{\otherground@semkitfreq}{0.56}
\newcommand{\building@semkitfreq}{14.1}
\newcommand{\fence@semkitfreq}{3.90}
\newcommand{\vegetation@semkitfreq}{39.3}
\newcommand{\trunk@semkitfreq}{0.51}
\newcommand{\terrain@semkitfreq}{9.17}
\newcommand{\pole@semkitfreq}{0.29}
\newcommand{\trafficsign@semkitfreq}{0.08}
\newcommand{\semkitfreq}[1]{{\csname #1@semkitfreq\endcsname}}

\newcommand{\car@sscbkitfreq}{2.85}
\newcommand{\bicycle@sscbkitfreq}{0.01}
\newcommand{\motorcycle@sscbkitfreq}{0.01}
\newcommand{\truck@sscbkitfreq}{0.16}
\newcommand{\othervehicle@sscbkitfreq}{5.75}
\newcommand{\person@sscbkitfreq}{0.02}
\newcommand{\road@sscbkitfreq}{14.98}
\newcommand{\parking@sscbkitfreq}{2.31}
\newcommand{\sidewalk@sscbkitfreq}{6.43}
\newcommand{\otherground@sscbkitfreq}{2.05}
\newcommand{\building@sscbkitfreq}{15.67}
\newcommand{\fence@sscbkitfreq}{0.96}
\newcommand{\vegetation@sscbkitfreq}{41.99}
\newcommand{\terrain@sscbkitfreq}{7.10}
\newcommand{\pole@sscbkitfreq}{0.22}
\newcommand{\trafficsign@sscbkitfreq}{0.06}
\newcommand{\otherstructure@sscbkitfreq}{4.33}
\newcommand{\otherobject@sscbkitfreq}{0.28}
\newcommand{\sscbkitfreq}[1]{{\csname #1@sscbkitfreq\endcsname}}

\usepackage{algorithm}
\usepackage{algorithmic}

\usepackage{newfloat}
\usepackage{listings}
\DeclareCaptionStyle{ruled}{labelfont=normalfont,labelsep=colon,strut=off} 
\floatstyle{ruled}
\newfloat{listing}{tb}{lst}{}
\floatname{listing}{Listing}
\title{Skip Mamba Diffusion for Monocular 3D Semantic Scene Completion}
\author{
    Li Liang \textsuperscript{\rm 1},
    Naveed Akhtar \textsuperscript{\rm 2},
    Jordan Vice \textsuperscript{\rm 1},
    Xiangrui Kong \textsuperscript{\rm 1},
    Ajmal Saeed Mian \textsuperscript{\rm 1}
}

\affiliations{
    \textsuperscript{\rm 1}The University of Western Australia \\
    \textsuperscript{\rm 2}The University of Melbourne \\
    li.liang@research.uwa.edu.au, naveed.akhtar1@unimelb.edu.au, jordan.vice@uwa.edu.au, xiangrui.kong@research.uwa.edu.au, ajmal.mian@uwa.edu.au

}

\usepackage{bibentry}

\begin{document}

\maketitle

\begin{abstract}
3D semantic scene completion is critical for multiple downstream tasks in autonomous systems. It estimates missing geometric and semantic information in the acquired scene data. Due to the challenging real-world conditions, this task usually demands complex models that process multi-modal data to achieve acceptable performance. We propose a unique neural model, leveraging advances from the state space and diffusion generative modeling to achieve remarkable 3D semantic scene completion performance with monocular image input. Our technique processes the data in the conditioned latent space of a variational autoencoder where diffusion modeling is carried out with an innovative state space technique. A key component of our neural network is the proposed Skimba (\textbf{Ski}p Ma\textbf{mba}) denoiser, which is adept at efficiently processing long-sequence data. The Skimba diffusion model is integral to our 3D scene completion network, incorporating a triple Mamba structure, dimensional decomposition residuals and varying dilations along three directions. We also adopt a variant of this network for the subsequent semantic segmentation stage of our method. Extensive evaluation on the standard SemanticKITTI and SSCBench-KITTI360 datasets show that our approach not only outperforms other monocular techniques by a large margin, it also achieves competitive performance against stereo methods. 
\begin{links}
\link{Code}{https://github.com/xrkong/skimba}
\end{links}

\end{abstract}
%

\section{Introduction}
3D semantic scene completion  is essential for inferring missing geometric and semantic information in scene data acquisition. This task finds numerous downstream applications in  autonomous driving \cite{li2022bevformer, hu2023planning}, robotic navigation \cite{tian2024occ3d, jin2024tod3cap}, and planning \cite{mei2023camera} etc. Recent semantic scene completion techniques use different input modalities, e.g., LiDAR, images, multi-modal, to pursue an acceptable level of performance. However, the complexity of the underlying modeling objective still makes 3D semantic scene completion an open challenge for the research community.

Several studies \cite{zhang2018efficient, roldao2020lmscnet, yan2021sparse, xiong2023learning, xu2023casfusionnet, xia2023scpnet} rely on LiDAR point clouds for 3D semantic scene completion. However, LiDAR data is inherently sparse - a property conflicting with the scene completion objective. Consequently, LiDAR-based scene completion models need to make up for this disparity with additional model complexity. For instance, \citet{xia2023scpnet} developed SCPNet, which enhances single-frame models through dense-to-sparse knowledge distillation. However, the model computational demands become sizable. Image-based methods \textcolor{black}{\cite{cao2022monoscene, zhang2023occformer, yu2024context, zheng2024monoocc}} utilize color images for the task, promising relatively simpler solutions. However, performance level of these techniques remains a bottleneck~\cite{zhang2023occformer, yu2024context}. As an example, \citet{yu2024context} introduced CGFormer, which utilizes multiple representations to encode 3D volumes from both local and global perspectives. Their method also  includes a depth refinement module to enhance depth estimation accuracy. Nevertheless, the advancement still lacks an acceptable performance for the critical nature of the task. 

Multi-modal inputs are also being considered for 3D semantic scene completion~\cite{li2019depth, li2020attention, cai2021semantic, dourado2022data, wang2023semantic, cao2024slcf}. As a representative example, FFNet~\cite{wang2022ffnet}  aims at addressing inconsistencies in RGB+D data and uncertainties in depth measurements. The technique employs a frequency fusion module and a correlation descriptor to capture the explicit correlation of RGB+D features. Whereas using multiple modalities helps in performance gain, it comes with a considerable computational overhead and intricacies in modeling. Moreover, achieving accurate alignment between different modalities remains a significant challenge along this direction.

\begin{figure*}[t] 
\centering   
\includegraphics[width=0.9\textwidth]{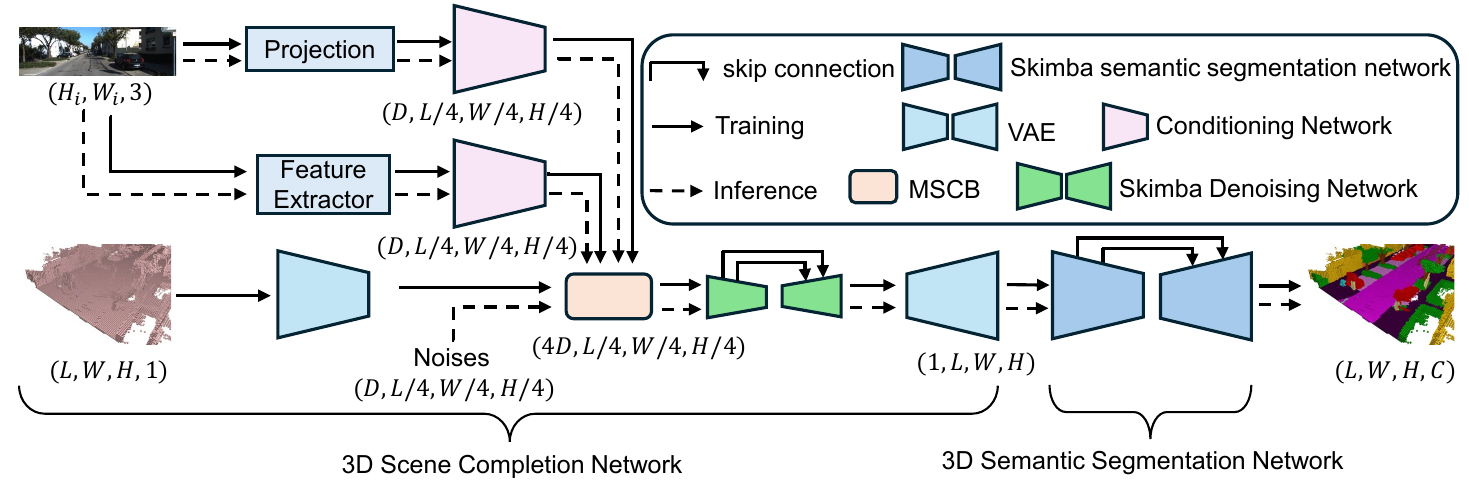}
\caption{Schematics of the approach. Our method comprises a 3D scene completion and a 3D semantic segmentation network. The former is encapsulated in a VAE framework that employs two sub-networks for conditioning its latent space, a Muti-Scale Convolutonal Block (MSCB) and a Skimba denoising network. The 3D semantic segmentation network employs a variant of Skimba. L, W, and H denote the length, width, and height of the original scene, and D is feature map dimension.}
\label{Method_1}
\end{figure*}

In this work, we focus on monocular image data to devise an effective 3D semantic scene completion method that demonstrates a marked performance gain in this category. Our approach leverages a unique combination of advances in generative diffusion modeling~\cite{rombach2022high} and state space models~\cite{gu2023mamba} tailored to the problem at hand. We propose a neural network, see Fig.~\ref{Method_1}, that encapsulates data processing for 3D scene completion in a conditioned latent space of a variational autoencoder (VAE). The conditioning lifts 2D scene information to 3D for subsequent multi-scale processing that prepares data for a diffusion-based denoiser. To that end, we propose a \textbf{Ski}p Ma\textbf{mba} (Skimba) diffusion model that extracts direct and indirect feature correspondences to provide sufficient contextual information for 3D semantic scene completion. Our Skimba diffusion model leverages a Triple Mamba configuration inspired by \cite{yang2024vivim}, where the proposed Skip Triple Mamba computes feature dependencies along three directions with varied dilations. Our network pays particular attention to efficient processing by deploying Skimba in a VAE latent space and further enabling downsampling within Skimba. A variant of Skimba is deployed as a 3D semantic segmentation sub-network at a later stage in the overall model. Our technique achieves impressive performance using for monocular 3D semantic scene completion. Key contributions of the paper are as follows. 
\begin{itemize}
    \item We propose a unique neural network that combines state space modeling with generative diffusion modeling of a conditioned latent space of a VAE for effective monocular 3D semantic scene completion.
    \item We introduce a Skimba diffusion model, employing skip triple mamba layers with varying dilations in three directions and carefully designed semantic blocks. A variant of Skimba is also deployed for the semantic segmentation sub-task in our network.
    \item With extensive experiments on SemanticKITTI and SSCBench-KITTI datasets, we demonstrate the state-of-the-art performance of our method. 
\end{itemize}

\section{Related Work}
Our contribution is in 3D semantic scene completion. For that, we devise a diffusion-based network that employs state space modeling. Hence, we organize  related work by discussing advances along the key topics of diffusion models, state space models and the 3D semantic scene completion.

\noindent{\bf Diffusion Model.} These models \cite{ho2020denoising, song2020score, song2020denoising} have been widely studied recently for generative tasks such as image generation \cite{zhou2023shifted, zheng2023layoutdiffusion}, image restoration \cite{zheng2024selective}, video generation \cite{ni2023conditional}, 3D shape generation \cite{shim2023diffusion, li2023diffusion}, and 3D occupancy prediction \cite{tang2024sparseocc}. Diffusion models are resource-intensive and require multiple function evaluations and gradient computations, leading to high costs \cite{patterson2021carbon}. To mitigate that, \citet{rombach2022high} introduced a latent diffusion model, which reduces computational demands by applying the diffusion process in the latent space of an auto-encoder. In 3D, while most research has focused on individual object generation,  \citet{tang2024sparseocc} has also tackled scene synthesis by employing a 3D sparse diffuser with spatially decomposed sparse kernels and a re-imagined sparse transformer head for 3D semantic occupancy prediction.

\noindent{\bf State Space Model.} State space modeling was originally developed in modern control theory \textcolor{black}{\cite{glasser1985control}} to describe dynamic systems, and has proven effective in modeling long-range dependencies. Recently, state space models have emerged as an alternative to convolutional neural networks (CNNs) and Transformers in natural language processing (NLP) and computer vision \cite{anthony2024blackmamba, li2024spikemba}. For example, S4 \cite{gu2021efficiently} shows impressive results in the vision domain using a diagonal parameter structure for efficient modeling. Subsequent models like HTTYH \cite{gu2022train}, DSS \cite{gupta2022diagonal}, and S4D \cite{gu2022parameterization} continue this trend with diagonal matrices, maintaining performance with reduced costs. S5 \textcolor{black}{\cite{smith2022simplified}} improves efficiency through parallel scanning, and MIMO SSM \cite{smith2022simplified} introduces mechanisms for linear-time inference and efficient training, further refining  Mamba \cite{gu2023mamba}. The Mamba architecture has been applied to image classification \cite{li2024mamba, zhu2024vision}, image segmentation \cite{anthony2024blackmamba}, and point clouds~\cite{liu2024point}. In particular, \cite{liu2024point} proposed Point Mamba, featuring an octree-based strategy for globally sorting raw points while preserving spatial proximity. 

\noindent{\bf 3D Semantic Scene Completion.} 3D semantic scene completion methods can be broadly divided into four main categories: image-based \cite{song2017semantic, yu2024context}, point cloud-based \cite{nie2021rfd, xiong2023learning}, voxel-based \cite{yan2021sparse, xia2023scpnet}, and multi-modality-based \cite{cai2021semantic, wang2022ffnet}. These methods commonly employ CNNs or transformers. For example, \citet{song2017semantic} proposed an end-to-end 3D convolutional network with a dilation-based context module to efficiently learn context with a large receptive field. \citet{xia2023scpnet} introduced SCPNet, which enhances single-frame model representations using dense relational semantic knowledge distillation and a label rectification technique to eliminate traces of dynamic objects. Although voxel-based methods are computationally more efficient than point-based methods, they often experience information loss. CNNs have limited receptive fields, and while transformers can expand receptive fields, they demand high memory usage. Mamba framework \cite{gu2023mamba} offers a balanced solution by expanding the receptive field without incurring high memory costs, which is promising for 3D semantic scene completion.

\section{Methodology}
The primary objective of monocular 3D semantic scene completion is to infer comprehensive 3D geometric and semantic information from a single image. Given an image $I \in \mathbb{R}^{L^{I} \times W^{I} \times{3}}$, it is projected to a grid of 3D voxels in $\mathbb R^{L \times W \times H}$. The task involves accurately completing the scene and assigning a class label to each voxel, determining whether it is empty or contains an object from one of $C$ semantic categories, represented as $c \in {0, 1, 2, \ldots, C-1}$.

\subsection{Proposed Network}
A schematic diagram of our proposed network is given in Fig.~\ref{Method_1}. A key component of our approach is the  \textbf{Ski}p ma\textbf{mba} (Skimba) denoising diffusion network. This proposed network forms an integral part of the overall technique such that its architecture is also leveraged for the 3D semantic segmentation required for the scene completion task. The overall approach employs a variational autoencoder (VAE) framework with two conditioning networks. These networks create lower-dimensional representations for the input voxel data and images, thereby reducing the computational demands without sacrificing performance. This notion is inspired by latent diffusion modeling~\cite{rombach2022high}. To the best of our knowledge, our approach provides the first successful demonstration of latent space diffusion for 3D scene completion tasks.  Our network employs a Multi-Scale Convolution Block (MSCB) to provide sufficient contextual information from conditioning features and noise.

\begin{figure*}[t] 
\centering   
\includegraphics[width=0.9\textwidth]{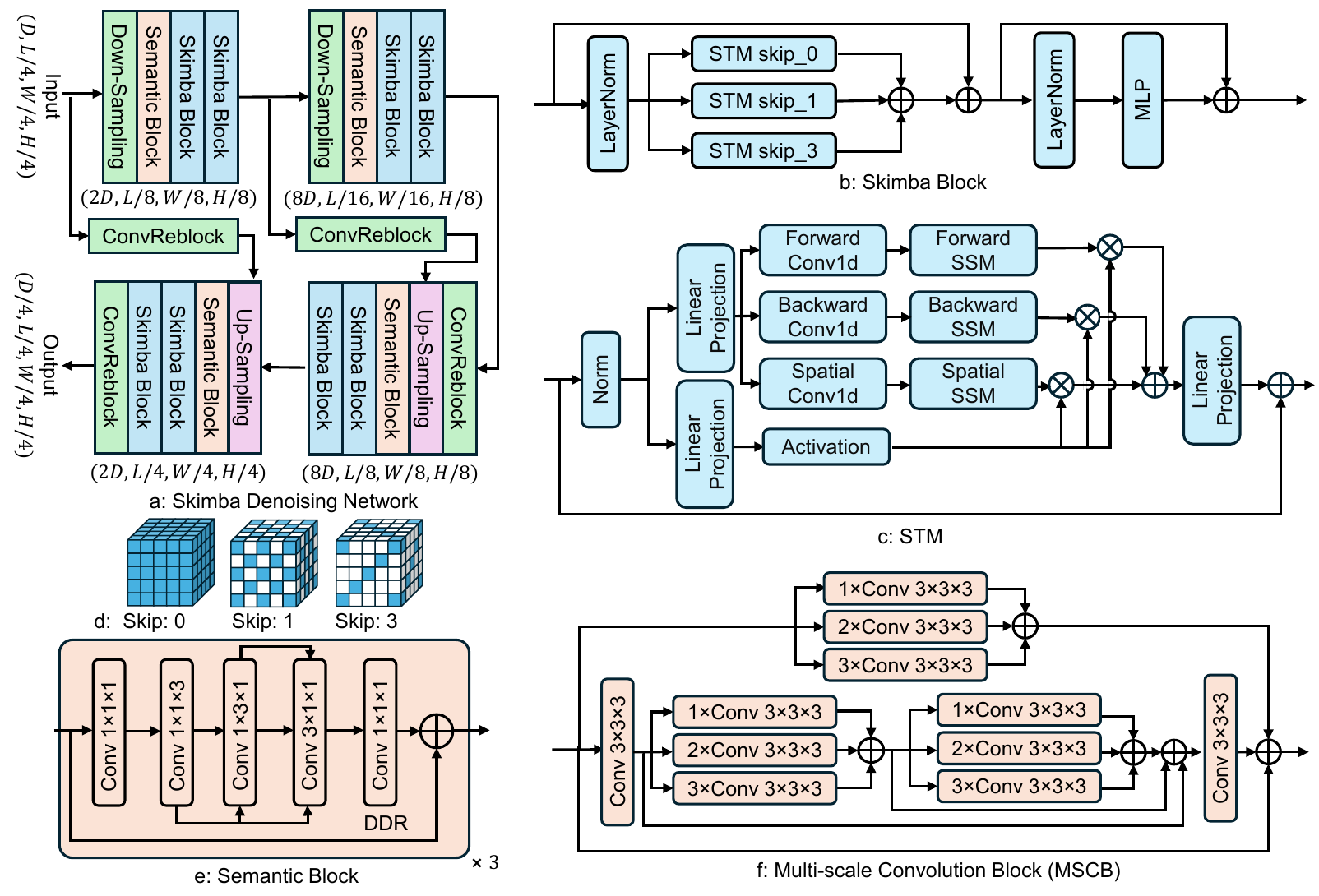}
\caption{Architectural details of the Skimba denoising network. Refer to the text for details.} 
\label{Method_2}
\end{figure*}

Below, we provide details of the constituent components of our technique. The  Skimba denoising network utilizes downsampling, semantic block (SB), Skimba block, ConvResblock, and upsampling block to achieve its objective. We discuss each of these components separately while providing relevant illustrations in Fig.~\ref{Method_2}.

\noindent{\bf Projection / Feature Extractor.} The use of Projection and Feature Extractor components is shown in Fig.~\ref{Method_1}. Their objective is to project  2D information to 3D, which is a challenging task due to the scale ambiguity present in a single viewpoint data available for our problem~\cite{fahim2021single}. The Projection component back-projects 2D features along their optical rays to create a unique 3D representation, enabling 2D-3D disentangled representations. This allows the subsequent 3D network to use high-level 2D features for detailed 3D disambiguation - a concept inspired by the approach of ~\citet{cao2022monoscene}. The Feature Extractor first converts image data into voxel representations through a projection layer. It then combines a multi-path block, proposed in SCPNet \cite{xia2023scpnet} with a Dimensional Decomposition Residual (DDR) block proposed by \citet{li2019rgbd}, stacking these blocks layer-by-layer. This enables the extraction of sufficient local and global contextual features for the subsequent condition network. 

\noindent{\bf Variational Autoencoder (VAE)/Condition Network.} 
As apparent from Fig.~\ref{Method_1}, our MSCB and Skimba denoising network operate in the latent space of a VAE that encapsulates the 3D scene completion network. To construct the VAE, we build on the insights of \cite{milletari2016v} and use an autoencoder trained with cross-entropy loss and Lovasz-softmax loss to ensure reconstructions remain on the grid manifold, avoiding the blurriness common with voxel-space losses like $L_2$ or $L_1$. For a given voxel scene $v \in L \times W \times H$, the encoder; say $E$, encodes $v$ into a latent representation $z = E(v)$. The decoder $D$ reconstructs it, yielding  $\tilde{v} = D(z) = D(E(v))$, such that  $z \in \mathbb{R}^{l \times w \times h \times c}$.   The encoder downsamples the voxel scene by a factor of $f = L/l = W/w = H/h$, where $f=4$ in our experiments. The diffusion model leverages the 3D structure of the learned latent space for efficient compression and high-quality reconstruction. The condition networks follow the encoder in their architecture to match their representation space with the VAE's latent space. 

\noindent{\bf \textcolor{black}{Multi-Scale Convolutional Block (MSCB).}} Our technique uses a Multi-Scale Convolution Block (MSCB) to efficiently extract multi-scale features. Details of this block are illustrated in Fig.~\ref{Method_2} (f). Our design is inspired by the multi-path block in SCPNet \cite{xia2023scpnet} and the popular  VGG architecture \cite{simonyan2014very}. The MSCB replaces larger convolution kernels with sequential \textcolor{black}{$3 \times 3 \times 3$} convolution layers. Two iterations replicate a \textcolor{black}{$5 \times 5 \times 5$} convolution, while three iterations approximate a \textcolor{black}{$7 \times 7\times 7$} convolution, significantly reducing computational costs. For example, the computational cost for a \textcolor{black}{$5 \times 5\times 5$} convolution is reduced from $125C^{2}$ to $54C^{2}$ for $C$ channels. For a \textcolor{black}{$7 \times 7\times 7$} convolution this reduction is from $343C^{2}$ to $81C^{2}$. This approach effectively captures both local and global contextual information while maintaining computational efficiency. In Fig.~\ref{Method_2}(f), we only show the $3\times 3 \times 3$ iteration for the MSCB. 

\noindent{\bf Skimba Denoising Network.} In the 3D Scene Completion Network, our Skimba denoising sub-network processes the MSCB output. This network is inspired by the Mamba framework~\cite{gu2023mamba}. Adopting state-space modeling, this framework maps a one-dimensional continuous function or sequence $x(t) \in \mathbb{R}$ to $y(t) \in \mathbb{R}$ through a hidden state $h(t) \in \mathbb{R}^{N}$. As a continuous-time framework, it requires discretizing the signal for deep learning applications to match the sampling rate of the input data. Ultimately, the model computes the output $y$ through a global convolution operation within a structured convolutional kernel. We refer to \cite{gu2023mamba} for details. Here, we explain the Skimba by discussing its constituent components. 

\noindent{\bf \textit{Down-sampling / Up-sampling / ConvReblock:}} 
In the Skimba network, the downsampling block, upsampling block, and ConvReblock are configured as shown in Fig.~\ref{Method_2}(a). The downsampling block comprises three sequential \textcolor{black}{$3 \times 3 \times 3$} convolution layers with instance normalization, where the first layer is followed by a LeakyReLU activation. A residual connection bypasses the first two layers and merges with the output of the third layer, followed by another LeakyReLU activation. This structure is conducive for effective gradient flow and facilitates learning complex representations. The ConvReblock shares this architecture but does not include downsampling because its purpose is to transmit residue from the encoding stage to the decoding stage of the underlying denoiser. The Up-sampling block starts with a transposed convolutional layer to increase the spatial resolution, followed by the same structure as the downsampling block, including a residual connection for effective feature reconstruction. Skimba uses multiple instances of all three types of blocks, as shown in Fig.~\ref{Method_2}(a). \\
\noindent{\bf \textit{Semantic Block:}} This is another type of block devised to help process the signal under the Mamba framework. The Mamba-based Skimba Block - discussed below - must flatten 3D features into a 1D sequence. Hence, we develop the Semantic Block (SB) to extract spatial relationships beforehand. The SB includes \textcolor{black}{Dimensional Decomposition Residual (DDR) \cite{li2019rgbd} blocks shown in Fig.~\ref{Method_2}(e) with different dilation rates to efficiently capture extensive scene layout information.} The DDR block decomposes 3D CNN computations, thereby reducing the computational cost. Traditional 3D CNN blocks have a computational cost of $C^{in} \times C^{out} \times k^{3}$, while the DDR block reduces this to $C^{in} \times C^{out} \times {3k}$ by breaking down the operations into $1 \times 1 \times k$, $1 \times k \times 1$, and $k \times 1 \times 1$ layers. This approach decreases the parameter count to one-third of what is required by a standard \textcolor{black}{$3 \times 3 \times 3$ convolution kernel} while maintaining detailed spatial layout.\\
\noindent{\bf \textit{Skimba Block:}} A key component of our Skimba denoiser is the Skimba block, shown in Fig.~\ref{Method_2}(b). This block follows a Triple Mamba (TM) configuration inspired by \cite{yang2024vivim} to capture both direct and indirect feature connections, thereby enriching the contextual information within high-dimensional features. We propose to use \textcolor{black}{Skip Triple Mamba (STM)} layers with varying dilations in three distinct directions: forward, reverse, and spatial. This is depicted in Fig.~\ref{Method_2}(c). Each STM layer computes feature dependencies using distinct dilation rates of 0, 1, and 3 in three directions, ensuring that comprehensive feature relationships are effectively captured. Fig.~\ref{Method_2}(d) show \textcolor{black}{the structure of the feature maps caused by the different dilation rates.} Let us denote the output of the $i^{\text{th}}$ STM layer by $\psi_i$, given as
\begin{align}
\psi_{i} = \psi_{i}^{f}(z) + \psi_{i}^{r}(z) + \psi_{i}^{s}(z),
\end{align}
where $z$ denotes the input features, and $f$, $r$, and $s$ are respectively the forward, reverse, and spatial directions. The fused 3D features from  different dilations are  expressed as
\begin{align}
\psi_{all} &= \sum_{i} \psi_{i}, i \in \left \{0, 1, 3 \right \},
\end{align}
where $\psi_{all}$ represents the combined features from the three STM layers. The final output feature  from the Skimba block; say $\phi_{all}$, can be expressed as follows
\begin{align}
\phi_{in} &= \psi_{all}\left(\text{LN}(f_{\text{initial}})\right) + f_{\text{initial}},\\
\phi_{all} &= \text{MLP}\left(\text{LN}\left(\phi_{in}\right)\right) + \phi_{in},
\end{align}
where $f_{\text{initial}}$ represents the input features of the Skimba block, while MLP and LN refer to stacked Linear layers and LayerNorm, respectively. By employing different dilations across three directions, the Skimba block effectively captures a wide range of feature dependencies, enhancing connectivity and improving performance in 3D semantic scene completion.
\textcolor{black}{The main difference between  Skimba block and   Mamba block \cite{li2024mamba, zhu2024vision} is that Skimba is optimized for memory-efficient extraction of rich direct and indirect spatial features by three STM layers, whereas Mamba captures spatial information from various scan directions, requiring considerable memory.}

\noindent{\bf \textcolor{black}{Skimba Semantic Segmentation Network.}} \textcolor{black}{The overall 3D semantic scene completion task can be divided into two sub-tasks: scene completion and semantic segmentation. In our approach, the latter is handled by the Skimba semantic segmentation network. The semantic segmentation network also follows an encoder-decoder architecture, as depicted in Fig.~\ref{Method_1}. This architecture shares similarities with the Skimba denoising diffusion network, with the primary distinction being the inclusion of a Skimba block only before the feature is fed into the decoder, without additional Skimba blocks after each SB. Furthermore, the skip connections are integrated throughout the architecture, linking the encoder and decoder stages. These connections allow the network to reuse feature maps from earlier layers, thereby preserving the spatial information and enhancing segmentation accuracy. This framework strikes an effective balance between efficiency and performance, enabling the network to process complex scenes with varying object scales and contexts.}

\subsection{Training Objective}
Due to the complex underlying task, our training objective is of composite nature. The objective function for the Skimba denoising network follows from the diffusion denoising framework, which minimizes the Expected squared error in the predicted output, defined as
\begin{align}
    L_{DM} = \mathbb{E}_{x,\epsilon \sim N\left ( 0, 1 \right ) ,t}\left [ \left \| \epsilon - \epsilon _{\theta}\left ( x_{t}, t  \right )   \right \| _{2}^{2}   \right ],
\end{align}
where $\epsilon _{\theta}\left ( x_{t}, t  \right )$ denotes an equally weighted sequence of denoising autoencoder, where $t = 1\dots T$. The model predicts a denoised variant of the input $x_{t}$, where $x_{t}$ itself is a noisy version of the input $x$ at time stamp $t$. For 3D the semantic segmentation, the  objective function is given as
\begin{align}
\mathcal{L} &= \mathcal{L}_{CE} + \beta \mathcal{L}_{Lovasz},
\end{align}
where $\beta$  is a balancing coefficient that adjusts the contribution of each loss component, $\mathcal{L}_{CE}$ is the  cross-entropy loss, and $\mathcal{L}_{Lovasz}$ is the Lovasz-softmax loss which optimizes mIoU - a crucial performance metric in semantic segmentation. The mathematical forms of both are as follows
\begin{align}
\mathcal{L}_{CE} &= - \sum_{i} y_{i} \log(\hat{y_{i}}),\\
\mathcal{L}_{Lovasz} &= \frac{1}{|C|} \sum_{c=1}^{C} J \left ( e(c) \right ),
\end{align}
where $\hat{y}_i$ and $y_{i}$ are the prediction and ground truth for the $i^{\text{th}}$ element of the  output. $J$ denotes the Lovasz extension of the IoU, a piecewise linear function that minimizes the mIoU error, and $e\left ( c  \right ) $ is the error vector for each class $c$ within the set of classes $C$. The objective function of the VAE is constructed by integrating $\mathcal{L}_{CE}$, $\mathcal{L}_{Lovasz}$, and the KL divergence regularization term.
\begin{table*}[!t]
 \normalsize 
	\newcommand{\clsname}[2]{
		\rotatebox{90}{
			\hspace{-6pt}
			\textcolor{#2}{$\blacksquare$}
			\hspace{-6pt}
			\renewcommand\arraystretch{0.6}
			\begin{tabular}{l}
				#1                                      \\
				\hspace{-4pt} ~\tiny(\semkitfreq{#2}\%) \\
			\end{tabular}
	}}
	\resizebox{\linewidth}{!}
	{
		\begin{tabular}{l|c|cc|cccccccccccccccccccc}
			\toprule		
			Method 								 & 
			Input 								 & 
			IoU 								 & 
			mIoU  								 &  
			\clsname{road}{road}                 & 
			\clsname{sidewalk}{sidewalk}         &        
			\clsname{parking}{parking}           & 
			\clsname{other-grnd.}{otherground}   & 
			\clsname{building}{building}         & 
			\clsname{car}{car} 					 & 
			\clsname{truck}{truck}               &
			\clsname{bicycle}{bicycle}           &
			\clsname{motorcycle}{motorcycle}     &
			\clsname{other-veh.}{othervehicle}   &
			\clsname{vegetation}{vegetation}     &
			\clsname{trunk}{trunk}               &
			\clsname{terrain}{terrain}           &
			\clsname{person}{person}             &
			\clsname{bicyclist}{bicyclist}       &
			\clsname{motorcyclist}{motorcyclist} &
			\clsname{fence}{fence}               &
			\clsname{pole}{pole}                 &
			\clsname{traf.-sign}{trafficsign}
			\\
			\midrule 
            LMSCNet~\cite{roldao2020lmscnet}     & Mono & 31.38          & 7.07           & 46.70          & 19.50          & 13.50          & 3.10           & 10.30          & 14.30          & 0.30          & 0.00          & 0.00          & 0.00          & 10.80          & 0.00           & 10.40          & 0.00          & 0.00          & 0.00          & 5.40           & 0.00          & 0.00          \\
            AICNet~\cite{li2020anisotropic}     & Mono   & 23.93          & 7.09           & 39.30          & 18.30          & 19.80          & 1.60           & 9.60           & 15.30          & 0.70          & 0.00          & 0.00          & 0.00          & 9.60           & 1.90           & 13.50          & 0.00          & 0.00          & 0.00          & 5.00           & 0.10          & 0.00          \\
            JS3C-Net~\cite{yan2021sparse}  &  Mono     & 34.00          & 8.97           & 47.30          & 21.70          & 19.90          & 2.80           & 12.70          & 20.10          & 0.80          & 0.00          & 0.00          & 4.10          & 14.20          & 3.10           & 12.40          & 0.00          & 0.20          & 0.20          & 8.70           & 1.90          & 0.30          \\
			MonoScene~\cite{cao2022monoscene} & Mono  & 34.16 & 11.08 & 54.70 & 27.10 & 24.80 & 5.70
			& 14.40 & 18.80 & 3.30 & 0.50 & 0.70 & 4.40  & 14.90 & 2.40  & 19.50 & 1.00  & 1.40
			& 0.40  & 11.10 & 3.30 & 2.10         \\
			TPVFormer~\cite{huang2023tri}        & Mono  &34.25 & 11.26 & 55.10 & 27.20 & 27.40 & 6.50
			& 14.80 & 19.20 & 3.70 & 1.00 & 0.50 & 2.30  & 13.90 & 2.60  & 20.40 & 1.10  & 2.40
			& 0.30  & 11.00 & 2.90 & 1.50 		  \\
			SurroundOcc~\cite{wei2023surroundocc}    & Mono  & 34.72 & 11.86 & \textbf{56.90} & 28.30 & 30.20 & 6.80 
			& 15.20 & 20.60 & 1.40 & 1.60 & 1.20 & \textbf{4.40}  & 14.90 & 3.40  & 19.30 & 1.40  & 2.00
			& 0.10  & 11.30 & 3.90 & 2.40         \\
			OccFormer~\cite{zhang2023occformer}        & Mono  & 34.53 & 12.32 & 55.90 & 30.30 & \textbf{31.50} & 6.50 & 15.70 & 21.60 & 1.20 & 1.50 & \textbf{1.70} & 3.20  & 16.80 & 3.90  & 21.30 & \textbf{2.20}  & 1.10
			& 0.20  & 11.90 & 3.80 & 3.70         \\
			IAMSSC~\cite{xiao2024instance}  & Mono  & 43.74 & 12.37 & 54.00 & 25.50 & 24.70 & 6.90 & 19.20 & 21.30 & 3.80 & 1.10 & 0.60 & 3.90 & 22.70 & 5.80 & 19.40 & 1.50 & 2.90 & \textbf{0.50} & 11.90 & 5.30 & 4.10 \\
            VoxFormer-S~\cite{li2023voxformer}        & Stereo  & 42.95 & 12.20 & 53.90 & 25.30 & 21.10 & 5.60
			& 19.80 & 20.80 & 3.50 & 1.00 & 0.70 & 3.70  & 22.40 & 7.50  & 21.30 & 1.40  & 2.60 
			& 0.20  & 11.10 & 5.10 & 4.90         \\
			VoxFormer-T~\cite{li2023voxformer}        & Stereo-T  & 43.21 & 13.41 & 54.10 & 26.90 & 25.10 & \textbf{7.30} & 23.50 & 21.70 & 3.60 & 1.90 & 1.60 & 4.10 & 24.40 & 8.10 & \textbf{24.20} & 1.60 & 1.10 & 0.00 & 13.10 & \textbf{6.60} & 5.70 \\
			DepthSSC~\cite{yao2023depthssc}           & Stereo  &  44.58 & 13.11 & 55.64 & 27.25 & 25.72 & 5.78
			& 20.46 & 21.94 & 3.74 & 1.35 & 0.98 & 4.17  & 23.37 & 7.64  & 21.56 & 1.34  & 2.79 & 0.28  & 12.94 & 5.87 & 6.23         \\
            HASSC-S~\cite{wang2024not}  & Stereo  & 43.40 & 13.34 & 54.60 & 27.70 & 23.80 & 6.20 & 21.10 & 22.80 & 4.70 & 1.60 & 1.00 & 3.90 & 23.80 & 8.50 & 23.30 & 1.60 & \textbf{4.00} & 0.30 & 13.10 & 5.80 & 5.50 \\
            H2GFormer-S~\cite{wang2024h2gformer}  & Stereo  & 44.20 & 13.72 & 56.40 & 28.60 & 26.50 & 4.90 & 22.80 & 23.40 & \textbf{4.80} & 0.80 & 0.90 & 4.10 & 24.60 & 9.10 & 23.80 & 1.20 & 2.50 & 0.10 & 13.30 & 6.40 & \textbf{6.30} \\
            \hline
            SkimbaDiff (ours)				  & Mono      & \textbf{46.95} & \textbf{13.78} & 55.70 &  \textbf{30.60} & 20.40 & 5.50  & \textbf{25.40} & \textbf{26.10} & 2.30 & \textbf{2.80} & 0.80 & 3.10 & \textbf{28.50} & \textbf{10.40} & 23.40 & 1.10 & 0.30 & 0.00 &  \textbf{17.30} & 5.20 & 3.00 \\
			\bottomrule
		\end{tabular}
	}
	\setlength{\abovecaptionskip}{0cm}
	\setlength{\belowcaptionskip}{0cm}
        \centering
        \caption{Quantitative results on the SemanticKITTI test set. The best results are highlighted in \textbf{bold}. Mono, Stereo, and Stereo-T refer to monocular, stereo, and temporal stereo-based methods.}
	\label{tab:sem_kitti_test}
\end{table*}
\begin{table*}[!t]
 \normalsize
	\newcommand{\clsname}[2]{
		\rotatebox{90}{
			\hspace{-6pt}
			\textcolor{#2}{$\blacksquare$}
			\hspace{-6pt}
			\renewcommand\arraystretch{0.6}
			\begin{tabular}{l}
				#1                                       \\
				\hspace{-4pt} ~\tiny(\sscbkitfreq{#2}\%) \\
			\end{tabular}
	}}
	\resizebox{\linewidth}{!}
	{
		\begin{tabular}{l|c|cc|cccccccccccccccccc}
			\toprule
			Method                            			&
			Input                                       &
			IoU                                 		&
			mIoU                                		&
			\clsname{car}{car}                      	&
			\clsname{bicycle}{bicycle}              	&
			\clsname{motorcycle}{motorcycle}        	&
		    \clsname{truck}{truck}                   	&
			\clsname{other-veh.}{othervehicle}      	&
			\clsname{person}{person}                	&
			\clsname{road}{road}                    	&
			\clsname{parking}{parking}              	&
			\clsname{sidewalk}{sidewalk}            	&
			\clsname{other-grnd.}{otherground}      	&
			\clsname{building}{building}            	&
			\clsname{fence}{fence}                  	&
			\clsname{vegetation}{vegetation}        	&
			\clsname{terrain}{terrain}              	&
			\clsname{pole}{pole}                    	&
			\clsname{traf.-sign}{trafficsign}       	&
			\clsname{other-struct.}{otherstructure} 	&
			\clsname{other-obj.}{otherobject}
			\\
			\midrule
			MonoScene~\cite{cao2022monoscene}  & Mono & 37.87 & 12.31 & 19.34 & 0.43 & 0.58 & 8.02  & 2.03 & 0.86 & 48.35        
			& 11.38 & 28.13 & 3.32 & 32.89 & 3.53 & 26.15 & 16.75 & 6.92 & 5.67 & 4.20 & 3.09  \\
			TPVFormer~\cite{huang2023tri}  & Mono  & 40.22 & 13.64 & 21.56 & 1.09 & 1.37 & 8.06  & 2.57 & 2.38 & 52.99        
			& 11.99 & 31.07 & 3.78 & 34.83 & 4.80 & 30.08 & 17.52 & 7.46 & 5.86 & 5.48 & 2.70  \\
			OccFormer~\cite{zhang2023occformer}  & Mono & 40.27 & 13.81 & 22.58 & 0.66 & 0.26 & 9.89  & 3.82 & 2.77 & 54.30       
			& 13.44 & 31.53 & 3.55 & 36.42 & 4.80 & 31.00 & 19.51 & 7.77 & 8.51 & 6.95 & 4.60  \\
			IAMSSC~\cite{xiao2024instance}  & Mono & 41.80 & 12.97 & 18.53 & 2.45 & 1.76 & 5.12 & 3.92 & 3.09 & 47.55 & 10.56 & 28.35 & 4.12 & 31.53 & 6.28 & 29.17 & 15.24 & 8.29 & 7.01 & 6.35 & 4.19 \\
                VoxFormer~\cite{li2023voxformer}  & Stereo  & 38.76 & 11.91 & 17.84 & 1.16 & 0.89 & 4.56  & 2.06 & 1.63 & 47.01 & 9.67  & 27.21 & 2.89 & 31.18 & 4.97 & 28.99 & 14.69 & 6.51 & 6.92 & 3.79 & 2.43  \\
			DepthSSC~\cite{yao2023depthssc}    & Stereo & 40.85 & 14.28 & 21.90 & 2.36 & 4.30 & 11.51 & 4.56 & 2.92 & 50.88
			& 12.89 & 30.27 & 2.49 & 37.33 & 5.22 & 29.61 & 21.59 & 5.97 & 7.71 & 5.24 & 3.51  \\
               \hline
            SkimbaDiff (ours) & Mono & 41.92 & 14.40 & 20.35 & 2.74 & 1.05 & 10.64 & 3.64 & 1.65 & 47.58 & 10.35 & 33.18 & 3.46 & 37.67 & 9.68 & 31.89 & 20.75 & 6.95 & 6.75 & 6.83 & 4.07 \\
			\bottomrule
		\end{tabular}
	}
	\setlength{\abovecaptionskip}{0cm}
	\setlength{\belowcaptionskip}{0cm}
        \centering
	\caption{Quantitative results on SSCBench-KITTI360 test set.  Mono and stereo refer to monocular and stereo methods.}
	\label{tab:kitti_360_test}
\end{table*}
\section{Experiments}
The proposed network is evaluated on two standard benchmarks for semantic scene completion; namely, the SemanticKITTI dataset \cite{behley2019semantickitti} and SSCBench-KITTI-360 \cite{li2023sscbench}. We also perform ablation experiments  to extensively evaluate the impact of individual components of our approach. \\
\noindent \textbf{{Datasets.}} The SemanticKITTI benchmark \cite{behley2019semantickitti} provides densely annotated urban driving scenes derived from the KITTI Odometry Benchmark \cite{geiger2012we}. This dataset consists of voxelized scenes within a spatial volume of $51.2m \times 51.2m \times 64m$, resulting in voxel grids of $256 \times 256 \times 32$ with a voxel size of 0.2m. It includes 10 sequences for training, 1 for validation, and 11 for testing, encompassing 20 semantic classes. Our approach uses RGB images with  $1220 \times 370$ resolution, cropped from the original resolution of $1226 \times 370$. The SSCBench-KITTI-360 dataset \cite{li2023sscbench} includes 7 sequences designated for training, 1 sequence for validation, and 1 sequence for testing, encompassing 19 semantic classes. The RGB images used as inputs have a resolution of $1408 \times 376$ pixels. We report results on popular standard metrics for scene completion and semantic segmentation on these datasets with standard evaluation practices.\\
\noindent \textbf{Evaluation metrics.} To assess our  framework and compare it with existing works, we use established evaluation metrics for semantic scene completion as outlined by Song \emph{et al.} \cite{guedes2018semantic}. Our evaluation focuses on two primary aspects: accurate geometric reconstruction of the scene and precise semantic segmentation of each voxel. We use Intersection over Union (IoU) to measure geometric completion performance, defined as $IoU_{i} = \frac{TP_{i}}{TP_{i} + FN_{i} + FP_{i}}$, where $TP_i$, $FP_i$, and $FN_i$ represent true positives, false positives, and false negatives for the $i^{\text{th}}$ class, respectively. Mean Intersection over Union (mIoU) is used to evaluate semantic accuracy across all categories, defined as $mIoU = \frac{1}{C} \sum_{c=1}^{C} IoU_{c}$, where $C$ is the total number of classes.

\subsection{Implementation Details}
\textcolor{black}{Our experiments were conducted using a single NVIDIA GeForce 4090 GPU with 24GB  RAM. The training memory required for each network is approximately 18GB.}
The VAE was trained for 24 epochs using the AdamW optimizer with an initial learning rate of 3e-4. The Skimba denoiser network was trained for 43 epochs with AdamW at a 1e-3 learning rate and 1e-4 weight decay. The denoising step in the Skimba network was set to 100. The Skimba 3D semantic segmentation network used AdamW with a 5e-3 learning rate and 1e-4 weight decay. A WarmupCosineLR scheduler was applied across all training processes to gradually reduce the learning rate for optimal performance. 

\subsection{Results}
Our comparative analysis assesses the performance of our proposed method against several state-of-the-art techniques. These methods include LMSCNet \cite{roldao2020lmscnet}, AICNet \cite{li2020anisotropic}, JS3C-Net \cite{yan2021sparse}, MonoScene \cite{cao2022monoscene}, TPVFormer \cite{huang2023tri}, SurroundOcc \cite{wei2023surroundocc}, OccFormer \cite{zhang2023occformer}, IAMSSC \cite{xiao2024instance}, VoxFormer \cite{li2023voxformer}, DepthSSC \cite{yao2023depthssc}, HASSC-S \cite{wang2024not}, H2GFormer-S \cite{wang2024h2gformer}.
\begin{figure*}[t]
    \centering
    \begin{subfigure}{\textwidth} 
        \centering   
        \includegraphics[width=0.9\textwidth]{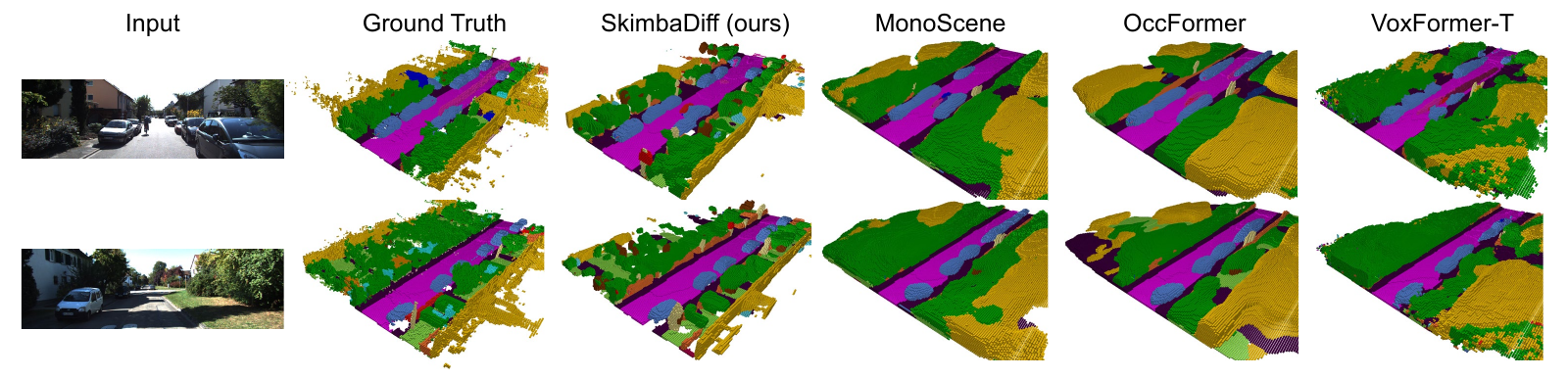}
    \end{subfigure}
    \vspace{0.5mm} 
    \begin{subfigure}{\textwidth} 
        \centering   
        \includegraphics[width=\textwidth]{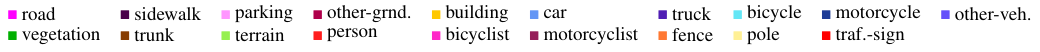}
    \end{subfigure}
    \caption{Qualitative results on the SemanticKITTI validation set. Columns from the left represent, input data, ground truth, and outputs of SkimbaDiff (our method), MonoScene, OccFormer, and VoxFormer-T (a stereo method).}
    \label{Comparison}
\end{figure*}

The results (obtained from the online server)  on SemanticKITTI test data for monocular 3D semantic scene completion are presented in Table~\ref{tab:sem_kitti_test}. These results highlight that our method achieves excellent performance. 
Note that, whereas our method only requires Monocular input, we also include results of the approaches relying on the more informative Stereo data. It is noteworthy that our method surpasses even some of the popular  Stereo methods. Specifically, the proposed method attains 46.95\% IoU and 13.82\% mIoU, outperforming the nearest competing monocular method by 3.21\% in IoU and 1.41\% in mIoU. This significant performance gain can be attributed to the exceptional capability of Skimba diffusion in completing and segmenting larger objects, such as sidewalks, buildings, vegetation, fences, and tree trunks. Additionally, the method demonstrates robust performance on smaller objects, including cars and bicycles, indicating its versatility across different object scales. \textcolor{black}{It is a common practice in the literature to report results on SemanticKITTI validation set as well. We include those results in the supplementary material, demonstrating equally strong performance.} 

In Table \ref{tab:kitti_360_test}, we summarize the results on the SSCBench-KITTI-360 test set. Again, our method outperforms recent monocular approaches and even some stereo-based 3D semantic scene completion methods. This consistent performance across different datasets underscores the robustness and generalizability of our approach. It is notable that this dataset has a smaller size and relatively lower sample quality, which adds to the complexity of the dataset. The dataset is more suited to methods that take LiDAR data as input, which show stronger performance on this dataset in general. Nevertheless, our method maintains its performance on this challenging dataset as well.

In Table \ref{tab:sem_time_table}, we present the inference time for each individual module within the proposed method and the inference time for the full model. Recall that the Skimba denoiser uses 100 steps. We can see that our Skimba denoising network takes a longer time compared to other models for this task. This is a common trade of generative diffusion models. Nevertheless, the proposed model offers much higher accuracy, which is desirable for a range of offline scene completion applications such as city planning, offline map completion for autonomous navigation etc. 

\begin{table}[t]
	\newcommand{\clsname}[2]{
		\rotatebox{90}{
			\hspace{-6pt}
			\textcolor{#2}{$\blacksquare$}
			\hspace{-6pt}
			\renewcommand\arraystretch{0.6}
			\begin{tabular}{l}
				#1                                      \\
				\hspace{-4pt} ~\tiny(\semkitfreq{#2}\%) \\
			\end{tabular}
	}}
	\centering
	\caption{Inference time (s) for each module. FE, CN, VAE, SD, SS, and FM are feature extractors,  conditioning networks, variational auto-encoder, the Skimba denoising network, the Skimba segmentation network, and full model.
 }
       \tiny
	\resizebox{\linewidth}{!}
	{    
		\begin{tabular}{l|l|l|l|l|l|r}
			\toprule		
			\multirow{2}{*}{Inference Time} & FE & CN & VAE & SD & SS & FD
                \\ \cline{2-7}
                \multirow{2}{*}{} & 0.15 & 0.21 & 0.13 & 7.38 & 0.26 & 8.13 \\
			\bottomrule
		\end{tabular}
	}
	\setlength{\abovecaptionskip}{0cm}
	\setlength{\belowcaptionskip}{0cm}
	\label{tab:sem_time_table}
\end{table}
\begin{table}[t]{
	\newcommand{\clsname}[2]{
		\rotatebox{90}{
			\hspace{-6pt}
			\textcolor{#2}{$\blacksquare$}
			\hspace{-6pt}
			\renewcommand\arraystretch{0.6}
			\begin{tabular}{l}
				#1                                      \\
				\hspace{-4pt} ~\tiny(\semkitfreq{#2}\%) \\
			\end{tabular}
	}}
	\centering
	\caption{Ablation study on SemanticKITTI validation set for 3D scene completion (IoU) and 3D semantic segmentation (mIoU). w/o denotes “without”.}
	\resizebox{\linewidth}{!}
	{    
		\begin{tabular}{l|c|cc}
			\toprule		
			\multirow{2}{*}{Method} 								 & 
			3D Scene Completion 								 &
                \multicolumn{2}{c}{3D Semantic Segmentation}                                   
			\\
                                                     &
                IoU                                  &
                IoU                                  &
                mIoU
                \\
			\midrule
             w/o MSCB  &  34.91 &36.43 & 10.47 \\
             w/o SB    &  39.87 &40.15 & 13.33 \\
             w/o Skimba & 37.58 & 41.46 & 11.81 \\
             Full Model &43.82 & 44.13 & 13.92 \\
			\bottomrule
		\end{tabular}
	}
	\setlength{\abovecaptionskip}{0cm}
	\setlength{\belowcaptionskip}{0cm}
	\label{tab:sem_kitti_val_ablation}}
\end{table}

We present representative visual results on the SemanticKITTI validation set in Fig.~\ref{Comparison}, where we used pre-trained models and the official implementations of MonoScene \cite{cao2022monoscene}, \textcolor{black}{OccFormer \cite{zhang2023occformer},} to generate their results. The figure also presents results of a Stereo method VoxFormer \cite{li2023voxformer} as a reference. These qualitative results demonstrate the excellent performance of our model, particularly in accurately segmenting planar categories such as sidewalks, buildings, vegetation, fences, and tree trunks. This visual evidence supports our quantitative findings, underscoring the effectiveness of our model in handling various categories.
\textcolor{black}{We present further qualitative result examples for our method in the supplementary material.} 

\subsection{Ablation Study}
We conducted systematic experiments to evaluate the impact of the various components in our framework, to quantify their contributions to the overall performance. The results of these ablation studies, detailed in Table \ref{tab:sem_kitti_val_ablation}, provide valuable insights into the importance of each component within the architecture. By systematically removing specific blocks, we were able to observe variations in the performance. This allows us to identify the critical elements that most significantly enhance the effectiveness of the model in the 3D semantic scene completion task. The findings from these experiments clearly indicate that both the MSCB and the Skimba are essential for achieving high performance. Their inclusion contributes to the ability of the model to accurately capture and represent complex spatial and semantic relationships. The without (w/o) Skimba refers to the triple mamba block, and does not involve the STM layer.

\section{Conclusion}
We proposed a 3D semantic scene completion network with a Skimba denoising diffusion sub-network. Our approach incorporates a variational autoencoder with two conditioning networks to produce lower-dimensional, perceptually equivalent symbolic spaces for input data, which effectively reduces computational demands while maintaining performance. The Mamba-inspired Skimba network captures both direct and indirect feature relationships within the data by using various skip triple dilations. This functionality enhances the ability of the network to represent the spatial and semantic structure of complex 3D scenes. Extensive evaluations on the SemanticKITTI and SSCBench-KITTI-360 datasets demonstrate that our method outperforms existing state-of-the-art methods, highlighting its effectiveness and potential for advancing 3D semantic scene completion.

\section{Acknowledgments}
Dr. Naveed Akhtar is a recipient of the ARC Discovery Early Career Researcher Award (project \# DE230101058), funded by the Australian Government. Professor Ajmal Mian is the recipient of an ARC Future Fellowship Award (project \# FT210100268) funded by the Australian Government.

\bibliography{aaai25}

\begin{thebibliography}{65}
\providecommand{\natexlab}[1]{#1}

\bibitem[{Anthony et~al.(2024)Anthony, Tokpanov, Glorioso, and Millidge}]{anthony2024blackmamba}
Anthony, Q.; Tokpanov, Y.; Glorioso, P.; and Millidge, B. 2024.
\newblock Blackmamba: Mixture of experts for state-space models.
\newblock \emph{arXiv preprint arXiv:2402.01771}.

\bibitem[{Behley et~al.(2019)Behley, Garbade, Milioto, Quenzel, Behnke, Stachniss, and Gall}]{behley2019semantickitti}
Behley, J.; Garbade, M.; Milioto, A.; Quenzel, J.; Behnke, S.; Stachniss, C.; and Gall, J. 2019.
\newblock Semantickitti: A dataset for semantic scene understanding of lidar sequences.
\newblock In \emph{ICCV}, 9297--9307.

\bibitem[{Cai et~al.(2021)Cai, Chen, Zhang, Lin, Wang, and Li}]{cai2021semantic}
Cai, Y.; Chen, X.; Zhang, C.; Lin, K.-Y.; Wang, X.; and Li, H. 2021.
\newblock Semantic scene completion via integrating instances and scene in-the-loop.
\newblock In \emph{CVPR}, 324--333.

\bibitem[{Cao and De~Charette(2022)}]{cao2022monoscene}
Cao, A.-Q.; and De~Charette, R. 2022.
\newblock Monoscene: Monocular 3d semantic scene completion.
\newblock In \emph{CVPR}, 3991--4001.

\bibitem[{Cao and Behnke(2024)}]{cao2024slcf}
Cao, H.; and Behnke, S. 2024.
\newblock SLCF-Net: Sequential LiDAR-Camera Fusion for Semantic Scene Completion using a 3D Recurrent U-Net.
\newblock \emph{arXiv preprint arXiv:2403.08885}.

\bibitem[{Dourado, Guth, and de~Campos(2022)}]{dourado2022data}
Dourado, A.; Guth, F.; and de~Campos, T. 2022.
\newblock Data augmented 3d semantic scene completion with 2d segmentation priors.
\newblock In \emph{WACV}, 3781--3790.

\bibitem[{Fahim, Amin, and Zarif(2021)}]{fahim2021single}
Fahim, G.; Amin, K.; and Zarif, S. 2021.
\newblock Single-View 3D reconstruction: A Survey of deep learning methods.
\newblock \emph{Elsevier: Computers \& Graphics}, 94: 164--190.

\bibitem[{Geiger, Lenz, and Urtasun(2012)}]{geiger2012we}
Geiger, A.; Lenz, P.; and Urtasun, R. 2012.
\newblock Are we ready for autonomous driving? the kitti vision benchmark suite.
\newblock In \emph{CVPR}, 3354--3361. IEEE.

\bibitem[{Glasser(1985)}]{glasser1985control}
Glasser, W. 1985.
\newblock \emph{Control theory}.
\newblock Harper and Row New York.

\bibitem[{Gu and Dao(2023)}]{gu2023mamba}
Gu, A.; and Dao, T. 2023.
\newblock Mamba: Linear-time sequence modeling with selective state spaces.
\newblock \emph{arXiv preprint arXiv:2312.00752}.

\bibitem[{Gu et~al.(2022{\natexlab{a}})Gu, Goel, Gupta, and R{\'e}}]{gu2022parameterization}
Gu, A.; Goel, K.; Gupta, A.; and R{\'e}, C. 2022{\natexlab{a}}.
\newblock On the parameterization and initialization of diagonal state space models.
\newblock \emph{NeurIPS}, 35: 35971--35983.

\bibitem[{Gu, Goel, and R{\'e}(2021)}]{gu2021efficiently}
Gu, A.; Goel, K.; and R{\'e}, C. 2021.
\newblock Efficiently modeling long sequences with structured state spaces.
\newblock \emph{arXiv preprint arXiv:2111.00396}.

\bibitem[{Gu et~al.(2022{\natexlab{b}})Gu, Johnson, Timalsina, Rudra, and R{\'e}}]{gu2022train}
Gu, A.; Johnson, I.; Timalsina, A.; Rudra, A.; and R{\'e}, C. 2022{\natexlab{b}}.
\newblock How to train your hippo: State space models with generalized orthogonal basis projections.
\newblock \emph{arXiv preprint arXiv:2206.12037}.

\bibitem[{Guedes, de~Campos, and Hilton(2018)}]{guedes2018semantic}
Guedes, A. B.~S.; de~Campos, T.~E.; and Hilton, A. 2018.
\newblock Semantic scene completion combining colour and depth: preliminary experiments.
\newblock \emph{arXiv preprint arXiv:1802.04735}.

\bibitem[{Gupta, Gu, and Berant(2022)}]{gupta2022diagonal}
Gupta, A.; Gu, A.; and Berant, J. 2022.
\newblock Diagonal state spaces are as effective as structured state spaces.
\newblock \emph{NeurIPS}, 35: 22982--22994.

\bibitem[{Ho, Jain, and Abbeel(2020)}]{ho2020denoising}
Ho, J.; Jain, A.; and Abbeel, P. 2020.
\newblock Denoising diffusion probabilistic models.
\newblock \emph{NeurIPS}, 33: 6840--6851.

\bibitem[{Hu et~al.(2023)Hu, Yang, Chen, Li, Sima, Zhu, Chai, Du, Lin, Wang et~al.}]{hu2023planning}
Hu, Y.; Yang, J.; Chen, L.; Li, K.; Sima, C.; Zhu, X.; Chai, S.; Du, S.; Lin, T.; Wang, W.; et~al. 2023.
\newblock Planning-oriented autonomous driving.
\newblock In \emph{CVPR}, 17853--17862.

\bibitem[{Huang et~al.(2023)Huang, Zheng, Zhang, Zhou, and Lu}]{huang2023tri}
Huang, Y.; Zheng, W.; Zhang, Y.; Zhou, J.; and Lu, J. 2023.
\newblock Tri-perspective view for vision-based 3d semantic occupancy prediction.
\newblock In \emph{CVPR}, 9223--9232.

\bibitem[{Jin et~al.(2024)Jin, Zheng, Li, Li, Zheng, Hu, Liu, Zhu, Yan, Sun et~al.}]{jin2024tod3cap}
Jin, B.; Zheng, Y.; Li, P.; Li, W.; Zheng, Y.; Hu, S.; Liu, X.; Zhu, J.; Yan, Z.; Sun, H.; et~al. 2024.
\newblock TOD3Cap: Towards 3D Dense Captioning in Outdoor Scenes.
\newblock \emph{arXiv preprint arXiv:2403.19589}.

\bibitem[{Li et~al.(2020{\natexlab{a}})Li, Han, Wang, Liu, and Yuan}]{li2020anisotropic}
Li, J.; Han, K.; Wang, P.; Liu, Y.; and Yuan, X. 2020{\natexlab{a}}.
\newblock Anisotropic convolutional networks for 3d semantic scene completion.
\newblock In \emph{CVPR}, 3351--3359.

\bibitem[{Li et~al.(2019{\natexlab{a}})Li, Liu, Gong, Shi, Yuan, Zhao, and Reid}]{li2019rgbd}
Li, J.; Liu, Y.; Gong, D.; Shi, Q.; Yuan, X.; Zhao, C.; and Reid, I. 2019{\natexlab{a}}.
\newblock Rgbd based dimensional decomposition residual network for 3d semantic scene completion.
\newblock In \emph{CVPR}, 7693--7702.

\bibitem[{Li et~al.(2019{\natexlab{b}})Li, Liu, Yuan, Zhao, Siegwart, Reid, and Cadena}]{li2019depth}
Li, J.; Liu, Y.; Yuan, X.; Zhao, C.; Siegwart, R.; Reid, I.; and Cadena, C. 2019{\natexlab{b}}.
\newblock Depth based semantic scene completion with position importance aware loss.
\newblock \emph{IEEE RA-L}, 5(1): 219--226.

\bibitem[{Li et~al.(2023{\natexlab{a}})Li, Duan, Zhou, and Lu}]{li2023diffusion}
Li, M.; Duan, Y.; Zhou, J.; and Lu, J. 2023{\natexlab{a}}.
\newblock Diffusion-sdf: Text-to-shape via voxelized diffusion.
\newblock In \emph{CVPR}, 12642--12651.

\bibitem[{Li, Singh, and Grover(2024)}]{li2024mamba}
Li, S.; Singh, H.; and Grover, A. 2024.
\newblock Mamba-nd: Selective state space modeling for multi-dimensional data.
\newblock \emph{arXiv preprint arXiv:2402.05892}.

\bibitem[{Li et~al.(2020{\natexlab{b}})Li, Zou, Li, Zhao, and Gao}]{li2020attention}
Li, S.; Zou, C.; Li, Y.; Zhao, X.; and Gao, Y. 2020{\natexlab{b}}.
\newblock Attention-based multi-modal fusion network for semantic scene completion.
\newblock In \emph{AAAI}, volume~34, 11402--11409.

\bibitem[{Li, Hong, and Fan(2024)}]{li2024spikemba}
Li, W.; Hong, X.; and Fan, X. 2024.
\newblock Spikemba: Multi-modal spiking saliency mamba for temporal video grounding.
\newblock \emph{arXiv preprint arXiv:2404.01174}.

\bibitem[{Li et~al.(2023{\natexlab{b}})Li, Li, Liu, Gong, Li, Chen, Wang, Li, Jiang, Yu et~al.}]{li2023sscbench}
Li, Y.; Li, S.; Liu, X.; Gong, M.; Li, K.; Chen, N.; Wang, Z.; Li, Z.; Jiang, T.; Yu, F.; et~al. 2023{\natexlab{b}}.
\newblock Sscbench: Monocular 3d semantic scene completion benchmark in street views.

\bibitem[{Li et~al.(2023{\natexlab{c}})Li, Yu, Choy, Xiao, Alvarez, Fidler, Feng, and Anandkumar}]{li2023voxformer}
Li, Y.; Yu, Z.; Choy, C.; Xiao, C.; Alvarez, J.~M.; Fidler, S.; Feng, C.; and Anandkumar, A. 2023{\natexlab{c}}.
\newblock Voxformer: Sparse voxel transformer for camera-based 3d semantic scene completion.
\newblock In \emph{CVPR}, 9087--9098.

\bibitem[{Li et~al.(2022)Li, Wang, Li, Xie, Sima, Lu, Qiao, and Dai}]{li2022bevformer}
Li, Z.; Wang, W.; Li, H.; Xie, E.; Sima, C.; Lu, T.; Qiao, Y.; and Dai, J. 2022.
\newblock Bevformer: Learning bird’s-eye-view representation from multi-camera images via spatiotemporal transformers.
\newblock In \emph{ECCV}, 1--18.

\bibitem[{Liu et~al.(2024)Liu, Yu, Wang, Zheng, Deng, Ye, and Wang}]{liu2024point}
Liu, J.; Yu, R.; Wang, Y.; Zheng, Y.; Deng, T.; Ye, W.; and Wang, H. 2024.
\newblock Point mamba: A novel point cloud backbone based on state space model with octree-based ordering strategy.
\newblock \emph{arXiv preprint arXiv:2403.06467}.

\bibitem[{Mei et~al.(2023)Mei, Yang, Wang, Zhu, Zhao, Ra, Li, and Liu}]{mei2023camera}
Mei, J.; Yang, Y.; Wang, M.; Zhu, J.; Zhao, X.; Ra, J.; Li, L.; and Liu, Y. 2023.
\newblock Camera-based 3D Semantic Scene Completion with Sparse Guidance Network.
\newblock \emph{arXiv preprint arXiv:2312.05752}.

\bibitem[{Milletari, Navab, and Ahmadi(2016)}]{milletari2016v}
Milletari, F.; Navab, N.; and Ahmadi, S.-A. 2016.
\newblock V-net: Fully convolutional neural networks for volumetric medical image segmentation.
\newblock In \emph{IEEE 3DV}, 565--571.

\bibitem[{Ni et~al.(2023)Ni, Shi, Li, Huang, and Min}]{ni2023conditional}
Ni, H.; Shi, C.; Li, K.; Huang, S.~X.; and Min, M.~R. 2023.
\newblock Conditional image-to-video generation with latent flow diffusion models.
\newblock In \emph{CVPR}, 18444--18455.

\bibitem[{Nie et~al.(2021)Nie, Hou, Han, and Nie{\ss}ner}]{nie2021rfd}
Nie, Y.; Hou, J.; Han, X.; and Nie{\ss}ner, M. 2021.
\newblock Rfd-net: Point scene understanding by semantic instance reconstruction.
\newblock In \emph{CVPR}, 4608--4618.

\bibitem[{Patterson et~al.(2021)Patterson, Gonzalez, Le, Liang, Munguia, Rothchild, So, Texier, and Dean}]{patterson2021carbon}
Patterson, D.; Gonzalez, J.; Le, Q.; Liang, C.; Munguia, L.-M.; Rothchild, D.; So, D.; Texier, M.; and Dean, J. 2021.
\newblock Carbon emissions and large neural network training.
\newblock \emph{arXiv preprint arXiv:2104.10350}.

\bibitem[{Roldao et~al.(2020)Roldao, de~Charette, Verroust-Blondet, Verroust-Blondet, and Verroust-Blondet}]{roldao2020lmscnet}
Roldao, L.; de~Charette, R.; Verroust-Blondet, R., Anne; Verroust-Blondet, R., Anne; and Verroust-Blondet, A. 2020.
\newblock Lmscnet: Lightweight multiscale 3d semantic completion.
\newblock In \emph{3DV}, 111--119.

\bibitem[{Rombach et~al.(2022)Rombach, Blattmann, Lorenz, Esser, and Ommer}]{rombach2022high}
Rombach, R.; Blattmann, A.; Lorenz, D.; Esser, P.; and Ommer, B. 2022.
\newblock High-resolution image synthesis with latent diffusion models.
\newblock In \emph{CVPR}, 10684--10695.

\bibitem[{Shim, Kang, and Joo(2023)}]{shim2023diffusion}
Shim, J.; Kang, C.; and Joo, K. 2023.
\newblock Diffusion-based signed distance fields for 3d shape generation.
\newblock In \emph{CVPR}, 20887--20897.

\bibitem[{Simonyan and Zisserman(2014)}]{simonyan2014very}
Simonyan, K.; and Zisserman, A. 2014.
\newblock Very deep convolutional networks for large-scale image recognition.
\newblock \emph{arXiv preprint arXiv:1409.1556}.

\bibitem[{Smith, Warrington, and Linderman(2022)}]{smith2022simplified}
Smith, J.~T.; Warrington, A.; and Linderman, S.~W. 2022.
\newblock Simplified state space layers for sequence modeling.
\newblock \emph{arXiv preprint arXiv:2208.04933}.

\bibitem[{Song, Meng, and Ermon(2020)}]{song2020denoising}
Song, J.; Meng, C.; and Ermon, S. 2020.
\newblock Denoising diffusion implicit models.
\newblock \emph{arXiv preprint arXiv:2010.02502}.

\bibitem[{Song et~al.(2017)Song, Yu, Zeng, Chang, Savva, and Funkhouser}]{song2017semantic}
Song, S.; Yu, F.; Zeng, A.; Chang, A.~X.; Savva, M.; and Funkhouser, T. 2017.
\newblock Semantic scene completion from a single depth image.
\newblock In \emph{CVPR}, 1746--1754.

\bibitem[{Song et~al.(2020)Song, Sohl-Dickstein, Kingma, Kumar, Ermon, and Poole}]{song2020score}
Song, Y.; Sohl-Dickstein, J.; Kingma, D.~P.; Kumar, A.; Ermon, S.; and Poole, B. 2020.
\newblock Score-based generative modeling through stochastic differential equations.
\newblock \emph{arXiv preprint arXiv:2011.13456}.

\bibitem[{Tang et~al.(2024)Tang, Wang, Wang, Zheng, Ren, Feng, and Ma}]{tang2024sparseocc}
Tang, P.; Wang, Z.; Wang, G.; Zheng, J.; Ren, X.; Feng, B.; and Ma, C. 2024.
\newblock Sparseocc: Rethinking sparse latent representation for vision-based semantic occupancy prediction.
\newblock In \emph{CVPR}, 15035--15044.

\bibitem[{Tian et~al.(2024)Tian, Jiang, Yun, Mao, Yang, Wang, Wang, and Zhao}]{tian2024occ3d}
Tian, X.; Jiang, T.; Yun, L.; Mao, Y.; Yang, H.; Wang, Y.; Wang, Y.; and Zhao, H. 2024.
\newblock Occ3d: A large-scale 3d occupancy prediction benchmark for autonomous driving.
\newblock \emph{NeurIPS}, 36.

\bibitem[{Wang et~al.(2023)Wang, Zhang, Zhang, Tang, and Sun}]{wang2023semantic}
Wang, F.; Zhang, D.; Zhang, H.; Tang, J.; and Sun, Q. 2023.
\newblock Semantic scene completion with cleaner self.
\newblock In \emph{CVPR}, 867--877.

\bibitem[{Wang et~al.(2024)Wang, Yu, Li, Liu, Liu, Chen, and Zhu}]{wang2024not}
Wang, S.; Yu, J.; Li, W.; Liu, W.; Liu, X.; Chen, J.; and Zhu, J. 2024.
\newblock Not all voxels are equal: Hardness-aware semantic scene completion with self-distillation.
\newblock In \emph{CVPR}, 14792--14801.

\bibitem[{Wang, Lin, and Wan(2022)}]{wang2022ffnet}
Wang, X.; Lin, D.; and Wan, L. 2022.
\newblock Ffnet: Frequency fusion network for semantic scene completion.
\newblock In \emph{AAAI}, volume~36, 2550--2557.

\bibitem[{Wang and Tong(2024)}]{wang2024h2gformer}
Wang, Y.; and Tong, C. 2024.
\newblock H2gformer: Horizontal-to-global voxel transformer for 3d semantic scene completion.
\newblock In \emph{AAAI}, volume~38, 5722--5730.

\bibitem[{Wei et~al.(2023)Wei, Zhao, Zheng, Zhu, Zhou, and Lu}]{wei2023surroundocc}
Wei, Y.; Zhao, L.; Zheng, W.; Zhu, Z.; Zhou, J.; and Lu, J. 2023.
\newblock Surroundocc: Multi-camera 3d occupancy prediction for autonomous driving.
\newblock In \emph{ICCV}, 21729--21740.

\bibitem[{Xia et~al.(2023)Xia, Liu, Li, Zhu, Ma, Li, Hou, and Qiao}]{xia2023scpnet}
Xia, Z.; Liu, Y.; Li, X.; Zhu, X.; Ma, Y.; Li, Y.; Hou, Y.; and Qiao, Y. 2023.
\newblock Scpnet: Semantic scene completion on point cloud.
\newblock In \emph{CVPR}, 17642--17651.

\bibitem[{Xiao et~al.(2024)Xiao, Xu, Kang, and Li}]{xiao2024instance}
Xiao, H.; Xu, H.; Kang, W.; and Li, Y. 2024.
\newblock Instance-aware monocular 3D semantic scene completion.
\newblock \emph{IEEE T-ITS}.

\bibitem[{Xiong et~al.(2023)Xiong, Ma, Wang, and Urtasun}]{xiong2023learning}
Xiong, Y.; Ma, W.-C.; Wang, J.; and Urtasun, R. 2023.
\newblock Learning compact representations for lidar completion and generation.
\newblock In \emph{CVPR}, 1074--1083.

\bibitem[{Xu et~al.(2023)Xu, Li, Tang, Yu, Hao, Hu, and Chen}]{xu2023casfusionnet}
Xu, J.; Li, X.; Tang, Y.; Yu, Q.; Hao, Y.; Hu, L.; and Chen, M. 2023.
\newblock Casfusionnet: A cascaded network for point cloud semantic scene completion by dense feature fusion.
\newblock In \emph{AAAI}, volume~37, 3018--3026.

\bibitem[{Yan et~al.(2021)Yan, Gao, Li, Zhang, Li, Huang, and Cui}]{yan2021sparse}
Yan, X.; Gao, J.; Li, J.; Zhang, R.; Li, Z.; Huang, R.; and Cui, S. 2021.
\newblock Sparse single sweep lidar point cloud segmentation via learning contextual shape priors from scene completion.
\newblock In \emph{AAAI}, volume~35, 3101--3109.

\bibitem[{Yang, Xing, and Zhu(2024)}]{yang2024vivim}
Yang, Y.; Xing, Z.; and Zhu, L. 2024.
\newblock Vivim: a video vision mamba for medical video object segmentation.
\newblock \emph{arXiv preprint arXiv:2401.14168}.

\bibitem[{Yao and Zhang(2023)}]{yao2023depthssc}
Yao, J.; and Zhang, J. 2023.
\newblock Depthssc: Depth-spatial alignment and dynamic voxel resolution for monocular 3d semantic scene completion.
\newblock \emph{arXiv preprint arXiv:2311.17084}.

\bibitem[{Yu et~al.(2024)Yu, Zhang, Ying, Yu, Hu, Luo, Cao, and Shen}]{yu2024context}
Yu, Z.; Zhang, R.; Ying, J.; Yu, J.; Hu, X.; Luo, L.; Cao, S.; and Shen, H. 2024.
\newblock Context and Geometry Aware Voxel Transformer for Semantic Scene Completion.
\newblock \emph{arXiv preprint arXiv:2405.13675}.

\bibitem[{Zhang et~al.(2018)Zhang, Zhao, Yao, Chen, Zhang, and Liao}]{zhang2018efficient}
Zhang, J.; Zhao, H.; Yao, A.; Chen, Y.; Zhang, L.; and Liao, H. 2018.
\newblock Efficient semantic scene completion network with spatial group convolution.
\newblock In \emph{ECCV}, 733--749.

\bibitem[{Zhang, Zhu, and Du(2023)}]{zhang2023occformer}
Zhang, Y.; Zhu, Z.; and Du, D. 2023.
\newblock Occformer: Dual-path transformer for vision-based 3d semantic occupancy prediction.
\newblock In \emph{ICCV}, 9433--9443.

\bibitem[{Zheng et~al.(2024{\natexlab{a}})Zheng, Wu, Yang, Zhang, Hu, and Zheng}]{zheng2024selective}
Zheng, D.; Wu, X.-M.; Yang, S.; Zhang, J.; Hu, J.-F.; and Zheng, W.-S. 2024{\natexlab{a}}.
\newblock Selective Hourglass Mapping for Universal Image Restoration Based on Diffusion Model.
\newblock In \emph{CVPR}, 25445--25455.

\bibitem[{Zheng et~al.(2023)Zheng, Zhou, Li, Qi, Shan, and Li}]{zheng2023layoutdiffusion}
Zheng, G.; Zhou, X.; Li, X.; Qi, Z.; Shan, Y.; and Li, X. 2023.
\newblock Layoutdiffusion: Controllable diffusion model for layout-to-image generation.
\newblock In \emph{CVPR}, 22490--22499.

\bibitem[{Zheng et~al.(2024{\natexlab{b}})Zheng, Li, Li, Zheng, Jin, Zhong, Long, Zhao, and Zhang}]{zheng2024monoocc}
Zheng, Y.; Li, X.; Li, P.; Zheng, Y.; Jin, B.; Zhong, C.; Long, X.; Zhao, H.; and Zhang, Q. 2024{\natexlab{b}}.
\newblock Monoocc: Digging into monocular semantic occupancy prediction.
\newblock \emph{arXiv preprint arXiv:2403.08766}.

\bibitem[{Zhou et~al.(2023)Zhou, Liu, Zhu, Yang, Chen, and Xu}]{zhou2023shifted}
Zhou, Y.; Liu, B.; Zhu, Y.; Yang, X.; Chen, C.; and Xu, J. 2023.
\newblock Shifted diffusion for text-to-image generation.
\newblock In \emph{CVPR}, 10157--10166.

\bibitem[{Zhu et~al.(2024)Zhu, Liao, Zhang, Wang, Liu, and Wang}]{zhu2024vision}
Zhu, L.; Liao, B.; Zhang, Q.; Wang, X.; Liu, W.; and Wang, X. 2024.
\newblock Vision mamba: Efficient visual representation learning with bidirectional state space model.
\newblock \emph{arXiv preprint arXiv:2401.09417}.

\end{thebibliography}

\end{document}